# KANNADA NAMED ENTITY RECOGNITION AND CLASSIFICATION (NERC) BASED ON MULTINOMIAL NAÏVE BAYES (MNB) CLASSIFIER

S Amarappa 1 & Dr. S V Sathyanarayana 2

Department of E & C Jawaharlal Nehru National College of Engineering,
Shimoga – 577 204, India

## ABSTRACT

Named Entity Recognition and Classification (NERC) is a process of identification of proper nouns in the text and classification of those nouns into certain predefined categories like person name, location, organization, date, and time etc. NERC in Kannada is an essential and challenging task. The aim of this work is to develop a novel model for NERC, based on Multinomial Naïve Bayes (MNB) Classifier. The Methodology adopted in this paper is based on feature extraction of training corpus, by using term frequency, inverse document frequency and fitting them to a tf-idf-vectorizer. The paper discusses the various issues in developing the proposed model. The details of implementation and performance evaluation are discussed. The experiments are conducted on a training corpus of size 95,170 tokens and test corpus of 5,000 tokens. It is observed that the model works with Precision, Recall and F1-measure of 83%, 79% and 81% respectively.

## 1. INTRODUCTION

India is rich with more than 1,652 mother tongues, out of which 22 are Scheduled Languages included in the Constitution of India. Among the 22 Scheduled Languages, Kannada is one of the major Dravidian languages of India, spoken predominantly in the state of Karnataka. The Karnataka Official Language Act 1963 recognized Kannada as its official language. The native speakers of Kannada are roughly 40 million, making it the 33rd most spoken language in the world. The language uses forty-nine phonemic letters, and the character set is almost identical to that of other Indian languages. Kannada is highly agglutinating and inflected language. It is a free word order language with rich heritage and large grammar. Processing of Kannada language and extraction of named entities is challenging. This language is inflected with three genders (masculine, feminine, and neutral) and two numbers (singular and plural). The Noun (ನಾಮಪಾಠ) is inflected by various factors such as case, number and gender.

Natural Language Processing (NLP) has two major tasks: Natural Language Understanding (NLU) and Natural Language Generation (NLG) Liddy (2001) [1]. NLU deals with machine reading comprehension i.e., the level of understanding of a text or message. NLG is the task of generating natural language from a machine representation system such as a knowledge base.

Apart from NLG and NLU, the other tasks to be done in NLP include automatic summarization, Information Extraction (IE), Information Retrieval (IR), Named Entity Recognition (NER) etc. In NLP, the primary goal of IE and IR is to automatically extract structured information. NER is a typical subtask of IE James Allen (2006) [2].

NERC involves processing of structured and unstructured documents and identifying proper names that refer to persons, organizations locations (cities, countries, rivers, etc), date, time etc. The aim of NERC is to automatically extract proper names which is useful to address many problems such as machine translation, information extraction, information retrieval, question answering, and automatic text summarization etc., Kavi N M (2006) [3].

In this paper, Kannada NERC based on MNB approach is dealt. The results obtained from the proposed model are quite encouraging with an average accuracy of 83%. What follows are the details of the proposed research work. Section 2 discusses about the details of existing work and challenges in current work. Section 3 deals with Naïve Bayes classifier principles, the technique used for NERC. Proposed methodology is dealt in section 4. Section 5 discusses implementation details. Finally the results are evaluated and discussed in section 6.

## 2. EXISTING WORK AND CHALLENGES IS CURRENT WORK

The NLP work was started way back in the 1940s. From 1940 to 1980, NLP systems were based on complex sets of hand-made rules. After 1980 NLP took new dimension, with machine learning algorithms. There is an enormous amount of data available for languages like English, but for Indian languages it is at the initial stage. Recent NLP algorithms are based on statistical machine learning. The term Named Entity was introduced in the sixth Message Understanding Conference (MUC-6), Gobinda Chowdhury (2003) [4]. The different techniques for addressing the NERC problem include: Maximum Entropy Models (Max-Ent) (Jaynes 1957), Hidden Markov Models (HMM) (Baum et al. 1966), Support Vector Machines (SVM) (Vapnik et al. 1992), Decision Trees (Sekine 1998) and Conditional Random Fields (CRF) (Lafferty et al. 2003) etc.

A lot of NLP work has been done in English, most of the European languages, some of the Asian languages like Chinese, Japanese, Korean and other foreign languages like Spanish, Arabic etc. NLP research in Indian languages is at the initial stage, as annotated corpora and other lexical resources have started appearing recently. In Computational Linguistics, Kannada is lagging far behind when compared to other Indian languages. In the following sections we are mentioning, a brief survey of research on NERC in Indian languages including Kannada. This is not a comprehensive and thorough survey, but is an indication of today's status in NERC research.

Benajiba et al. (2009) [5] discussed about SVM based Language Independent and Language Specific Features to Enhance Arabic NER. Padmaja et al. (2010) [6] discussed on the first steps towards Assamese NER. Asif Ekbal et al. (2008) [7] developed an NER system in Bengali using CRF approach. Ekbal and Shivaji (2008) [8] reported about the development of a NER system for Bengali using SVM. Ekbal and Shivaji (2008) [9] discussed about Bengali Named Entity Tagged Corpus and its use in NER Systems.

A Lot of work on NERC has been done in English language and here are quoted a few recent works. Maksim et al. (2012) [10] build a CRF based system that achieves 91.02% F1-measure on the CoNLL 2003 (Sang and Meulder, 2003) dataset and 81.4% F1-measure on the Onto Notes version 4 (Hovy et al., 2006) CNN dataset. Mansouri et al. (2008) [11] proposed a robust and novel Machine Learning based method called Fuzzy support Vector Machine (FSVM) for NER. Nadeau et al. (2006) [12] devised an unsupervised NER by Generating Gazetteers and Resolving

Ambiguity. Monica et al. (2009) [13] discussed about the evaluation of Named Entity Extraction Systems.

Sujan et al. (2008) [14] developed system for NER in Hindi using Max-Ent and Transliteration. In Li and McCallum (2004) [15] the authors have used CRF with feature induction to the Hindi NER task. Sujan et al. (2008) [16] developed a NER system for Hindi using Max-Ent. Sudha and Nusrat (2013) [17] experimented, NER using HMM on Hindi, Urdu and Marathi Languages. Deepti and Sudha (2013) [18] deviced algorithm for the Detection and Categorization of Named Entities in Indian Languages using HMM. Nusrat et al. (2012) [19] devised algorithm for NER in Indian Languages using Gazetteer method and HMM. S. Biswas et al. (2010) [20] developed a Two Stage Language Independent NER for Hindi, Oriya, Bengali and Telugu. Animesh et al. (2008) [21] talked about a new approach to recognize named entities for Indian languages. Sujan et al. (2008) [22] described a hybrid system that applies Max-Ent, language specific rules and gazetteers to the task of NER in Indian languages. Erik and Fien (2003) [23] gave Introduction to the CoNLL-2003 Shared Task a Language-Independent NER. Ekbal and Shivaji (2010) [24] reported about the development of a language independent NER system for Bengali and Hindi using SVM. Ekbal et al. (2008) [25] developed Language Independent NER system for South and South East Asian languages, particularly for Bengali, Hindi, Telugu, Oriya and Urdu as part of the IJCNLP-08 NER Shared Task1.

Kishorjit et al. (2011) [26] developed a model on CRF Based NER in Manipuri. Thoudam et al. (2009) [27] developed NER for Manipuri using SVM. Sitanath et al. (2010) [28] described a hybrid system that applies Max-Ent model with HMM and some linguistic rules to recognize Name Entities in Oriya language. Vishal and Gurpreet (2011) [29] explained about the NER System for Punjabi language text summarization using a Condition based approach. Pandian et al. (2008) [30] presented the construction of a hybrid, three stage NER for Tamil. Raju et al. (2008) [31] described a Max-Ent NER system for Telugu. Vijayanand and Seenivasan (2011) [32] devised NER and Transliteration for Telugu. Srikanth and KN Murthy (2008) [33] developed NER for Telugu using CRF based Noun Tagger. Praneeth et al. (2008) [34] conducted experiments in Telugu NER using CRF.

Shambhavi et al. (2012) [35] developed A Max-Ent model to Kannada Part Of Speech Tagging. Ramasamy et al. (2011) [36] proposed and developed a rule based Kannada Morphological Analyzer and Generator (MAG) using finite state transducer. Amarappa and Sathyanarayana [37] (2013) developed a HMM based system for NERC in Kannada Language.

Based on the survey, it is observed that a lot of work on NERC has been done in English and other foreign languages. NERC work in Indian languages is still in its initial stage. As for as Indian languages are concerned, some works related to NERC are found in Hindi, Bengali, Telugu, Tamil, Oriya, Manipuri, Punjabi, Marathi and Assamese Languages. In Kannada Language, some works on Kannada Morphology are reported in [35] [56]. In our earlier work [37] we have carried out NERC work in Kannada using HMM on a limited corpus of 10,000 tokens, however the works on NERC in Kannada are yet to be investigated and implemented. This motivated us to take up NERC in Kannada as the proposed Research area.

## 2.1 Challenges and Issues specific to Kannada language
Kannada language has no capitalization. It is Brahmi script with high phonetic characteristics that could be utilized by NERC system. There is non-availability of large gazetteer, lack of standardization and spelling. There are a number of frequently used words (common nouns), which can also be used as names. There is a lack of annotated data and it is highly agglutinating and inflected language.

## 3. MULTINOMIAL NAÏVE BAYES (MNB) CLASSIFIER

The Naïve Bayes classifier is a Generative Model of supervised learning algorithms. It is a simple probabilistic classifier which is based on Bayes' theorem with strong and naïve independence assumptions between every pair of features. It is one of the most basic classifier used for text classification. Moreover, the training time with Naïve Bayes is significantly smaller as opposed to alternative methods such as Support Vector Machine (SVM) and Maximum Entropy (Max-Ent) classifiers. Naïve Bayes classifier is superior in terms of CPU and memory consumption as shown by Huang, J. (2003). Its performance is very close to SVM and Max-Ent classifiers.

The Multinomial Naïve Bayes classifier is suitable for classification with discrete features. The multinomial distribution normally requires integer feature counts; however, fractional counts such as Term Frequency and Inverse Document Frequency (tf-idf) will also work. Multinomial Naïve Bayes classifier is based on the Naïve Bayes algorithm. In order to find the probability for a label, this algorithm uses the Bayes rule to express P (label | features) in terms of P (label) and P (features | label). The Naïve Bayes classifier requires training data samples in the format: (xi, yi) where, xi includes the contextual information of the word/document (the sparse array) and yi, its class. Graphical representation of Naïve Bayes decoder is shown in Figure1. Here fi is ith feature of vocabulary (vi = xi) and P (fi |yj) = P (xi = vi | yj) is the maximum probability that the input xi belongs to the class yj.

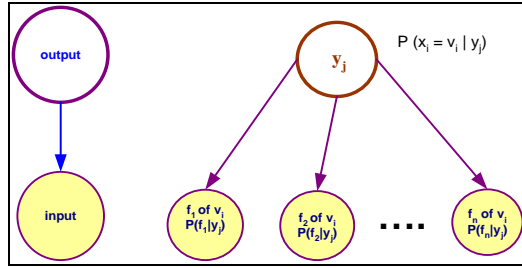

Figure 1. Graphical Model Of Naïve Bayes Decoding

Given a class variable y and a dependent feature vector x1 through xn, Bayes' theorem states the following relationship of Joint Probability:

$$P(X, Y) = P(Y, X)$$
$$P(X) \times P(Y \mid X) = P(Y) \times P(X \mid Y) \quad (1)$$

$$P(Y \mid X) = \frac{P(Y) \times P(X \mid Y)}{P(X)} \propto P(Y) \times P(X \mid Y) \quad (2)$$

$$P(X \mid Y) = \frac{P(X) \times P(Y \mid X)}{P(Y)} \propto P(X) \times P(Y \mid X) \quad (3)$$

Equation (2) can be alternatively written as:

$$P(y_j \mid x_1, x_2, ..., x_n) = \frac{P(y_j) \times P(x_1, x_2, ..., x_n \mid y_j)}{P(x_1, x_2, ..., x_n)} \quad (4)$$

Using the Naïve independence assumption $P(x_i \mid y_j, x_1, ...., x_{i-1}, x_{i+1}, ..., x_n) = P(x_i \mid y_j)$ for all i, Equation (4) is simplified to:

$$P(y_j \mid x_1, x_2, ..., x_n) = \frac{P(y_j) \times \prod_{i=1}^{n} P(x_i \mid y_j)}{P(x_1, x_2, ..., x_n)} \qquad (5)$$

Since $P(x_1, x_2, ..., x_n)$ is constant for given input, we can use the following classification rule:

$$P(y_j \mid x_1, x_2, ..., x_n) \propto P(y_j) \times \prod_{i=1}^{n} P(x_i \mid y_j)$$
$$\Downarrow \qquad (6)$$
$$y = \underset{y_j \in Y}{\arg\max} \prod_{i=1}^{n} P(x_i \mid y_j)$$

P (xi | yj ) is the relative frequency of class yj in the training set. $y$ is the Maximum probability of generating instance xi for the given class yj.

## 4. PROPOSED WORK AND METHODOLOGY

The main aim of this work is to develop a Supervised Statistical Machine Learning NERC system for Kannada Language based on MNB classifier. NERC involves identification of proper names in texts, and classification of those names into a set of pre-defined categories of interest such as: Person names (names of people), Organization names (companies, government organizations, committees, etc.), Location names (cities, countries etc.), and miscellaneous names (date, time, number, percentage, monetary expressions, number expressions and measurement expressions). The functional block diagram of the proposed system is as shown in Figure2. NERC in Kannada is important, because it gives solution to many applications of NLP such as web searching, scanning a set of documents written in a natural language and populating the database, building of useful dictionaries, constructing sophisticated word processors for Natural Languages, Information Extraction, Information Retrieval, Data mining, Publishing Books of Names, Places, Organizations etc. With the above context, the proposed system is designed.

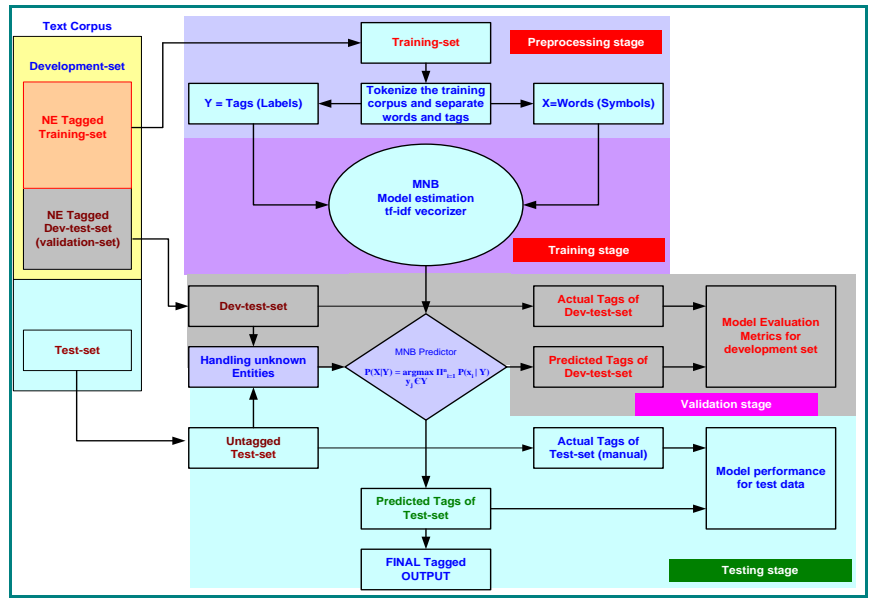

Figure 2. Named Entity Recognition and Classification (NERC) In Kannada Based On MNB

The design of proposed the system and methodology goes as follows:
1. Corpus creation and usage

Kannada NERC is very hard without tagged Corpora and hence we manually tagged about 100K Kannada words. This Kannada Corpus is used to build the NERC Model. The manually tagged Corpus include: Part of EMILLE (Enabling Minority Language Engineering) corpus, a part of the corpus taken from web articles and part of the corpus self typed from Kannada books. The whole corpus is divided into two sets: Development-set and Test-set as shown in Figure2. First, select the Development-set and then subdivide it into Training-set and development test set (Dev-test-set). The Training-set is used to train the model, and the Dev-test-set is used to perform error analysis. The Test-set serves for the final evaluation of the system. The machine learning used in the work is fully supervised MNB.

Take Training data from the development-set:

Input (training) data points : $X = [X_1, X_2, X_3, ..., X_M]$;

Or : $X = \text{array}(([x_{11}, x_{21}], [x_{12}, x_{22}], ... [x_{1M}, x_{2M}]))$

Labels (states) : $Y = [y_1, y_2, y_3, ..., y_N]$ ; $y_j = j^{th}$ label

2. Pre-processing stage

Here, the tagged text corpus is tokenized into words and tags (labels). The separated words are $X = [w_1, w_2, w_3, ..., w_M]$ and separated tags (labels) are $Y = [l_1, l_2, l_3, ..., l_N]$.

3. Training stage for the model

- From the separated words (X) and tags (Y) in pre-processing stage, extract the vocabulary (feature set) and unique labels. Let vocabulary be $V = [v_1, v_2, v_3, ..., v_k]$ or $F = [f_1, f_2, f_3, ..., f_k]$ and unique labels be $Y = [l_1, l_2, l_3, ..., l_j]$ (here M reduces to k and N reduces to j)

  - Find raw count of each vocabulary i.e., term frequencies (tf) according to their label.
  - The term-frequency is a measure of how many times a particular term of V(t), is present in the documents d1,d2, . dM. The term-frequency is defined as a counting function:

$$tf(t,d) = \sum_{x \in t} fr(x,t) \qquad (8)$$

where $fr(x,t)$ is a simple function, defined as:

$$fr(x,t) = \begin{cases} 1, & \text{if } x = t \\ 0, & \text{otherwise} \end{cases} \qquad (9)$$

The $tf(t,d)$ returns count of t in document d and it can be shown in matrix form:

$$M_{|D| \times X}(M_{train}) \qquad (10)$$

- Find inverse document frequency $idf(t)$ of training corpus defined by the function $p(t|d) = \frac{|\{d : t \in d\}|}{|D|}$, so *idf* is defined as:

$$idf = -\log p(t|d) = \log \frac{1}{p(t|d)} = \log \frac{|D|}{|\{d : t \in d\}|}$$

$$idf(t) = \ln\left(\frac{|D|+1}{1+|\{d:t \in d\}|}\right) + 1 \qquad (11)$$

Here $|\{d : t \in d\}|$ is the number of documents where the term *t* appears; when the term-frequency function satisfies $tf(t,d) \neq 0$. It should be noted that adding 1 into the formula above avoids zero division.

- Now to find *tf-idf* use the following steps:
$$tf - idf(t) = tf(t,d) \times idf(t) \tag{12}$$
  ➤ Find *idf* for each feature present in the feature matrix with the term frequency and *idf* weights can be represented by a vector as given by
$$\overrightarrow{idf}_{train} = [idf(t_1), idf(t_2), idf(t_3), ...., idf(t_k)] \tag{13}$$
  ➤ tf-idf matrix of training set in un-normalized form:
  Now the *tf* matrix, $M_{|D| \times X} = M_{train}$ of equation (10) and the *idf* matrix $\overrightarrow{idf}_{train} = M_{idf}$ of equation (13) are multiplied to calculate the *tf-idf* weights. (Among M documents *i* number of documents are taken for training)
  ➤ And then multiply $M_{idf}$ to the term frequency matrix, so the final result can be defined as:
$$[M_{tf-idf}]_{i \times k} = [M_{train}]_{i \times k} \times [M_{idf}]_{k \times k} \tag{14}$$
  ➤ tf-idf matrix of Training-set in normalized form:
$$M_{tf-idf} = \frac{M_{tf-idf}}{\|M_{tf-idf}\|_2} \tag{15}$$
  ➤ tf-idf vectors are the actual trained parameters of the MNB model **(Scikit-learn version 0.14 documentation).**

4. Validation stage:
Reserve a fold of the annotated training data for the Dev-test-set. Perform multiple evaluations on different Dev-test-sets and combine the scores from those evaluations. (http://www.nltk.org/book/ch06.html). Take a fold of the annotated training data as Dev-test-set from the Development-set and perform the following computations:
   - Pre-processing and tf-idf computation.
   - Compute the probability: $P(X|Y) = \underset{y_j \in Y}{\arg\max} \prod_{i=1}^{n} P(x_i|Y)$ as shown in Figure1.

5. Test(decoding) stage:
Take test data from the corpus set and do the following computations:
   - Pre-processing and tf-idf computation
   - Compute the probability: $P(X|Y) = \underset{y_j \in Y}{\arg\max} \prod_{i=1}^{n} P(x_i|Y)$

## 5. IMPLEMENTATION

The proposed system is designed based on MNB classification as discussed in section 4. The proposed model is as shown in Figure2. The system is programmed using Python 2.7 and Sklearn toolkit. The program is executed on windows platform using Intel Core2Duo CPU @3.00 GHz, a state of the art machine. The following are the various steps of implementation.

1. Kannada Baraha editor is used to manually create named entity tagged Kannada corpus in UTF-8 encoding format.
2. Twenty-two Named Entities (NEs) tabulated in Table1 are considered. A non-named entity is tagged as NONE.
3. From the tagged corpus, separate words, tags and store in separate lists. For the separated tags assign appropriate tag-labels.
4. Separated words and tag-labels are fed as Training-set to the MNB Model. In the training stage, MNB extracts features from the training corpus, such as vocabulary words, tf matrix, idf matrix and tf-idf matrix.
5. Test-set sequence is given as input to the model.
6. The features of the Test-set sequence are calculated and compared with the trained features. Accordingly each word is tagged with appropriate Named Entities (NEs).
7. Model performance is evaluated by finding Precession, Recall, and F1-measure.
8. Evaluated parameters are tabulated and plotted for better comparison.

Table 1. Named Entity Tag set.

| NE | Tag | Tag-label | Meaning | Example |
|---|---|---|---|---|
| Person | NEP | 0 | Name of a person one word | ಈಶ್ವರಚಂದ್ರ (ISvaracaMdra) |
| | NEPB | 13 | Beginning name | ಮೋಹನ್ (mOhan) |
| | NEPI | 14 | Intermediate name | ದಾಸ್ (dAs) |
| | NEPE | 15 | End name | ಗಾಂಧಿ (gAMdhi) |
| Location | NEL | 1 | Name of a place, location one word | ಕರ್ನಾಟಕ, ಶಿವಮೊಗ್ಗ (karnATaka, shivamogga) |
| | NELB | 16 | Beginning name Loc | ಯುನೈಟೆಡ್ (yunaiTeD) |
| | NELI | 17 | Intermediate name Loc | ಸ್ಟೇಟ್ಸ್ ಆಫ್ (sTETs Af) |
| | NELE | 18 | End name Loc | ಅಮೇರಿಕ (amErika) |
| Organization | NEO | 2 | Name of an organization one word | ನಗರಸಭೆ (nagarasabhe) |
| | NEOB | 19 | Beginning name Org | ಭಾರತೀಯ (bhAratIya) |
| | NEOI | 20 | Intermediate name Org | ವಿಜ್ಞಾನ (vij~jAna) |
| | NEOE | 21 | End name Org | ಸಂಸ್ಥೆ (saMsthe) |
| Designation | NED | 3 | Name of any designation | ಜೆನರಲ್ ಮ್ಯಾನೇಜರ್, ಕಮಿಷನರ್ (jenaral myAnEjar, kamiShnar) |
| Term | NETE | 4 | Name of terms, diseases | ಸಿದ್ಧಾಂತ, ನಿಯಮ, ಕಾಲರ (siddhAMta, niyama, kAlara) |
| Title-Person | NETP | 5 | Title before the name | ಡಾ||, ಶ್ರೀ, ಶ್ರೀಯುತ, ಮಹಾತ್ಮ (DA||,SrI,SrIyuta,mahAtma) |
| Title-Object | NETO | 6 | Name of Object | ಕುರ್ಚಿ, ಮೇಜು (kurchi, mEju) |
| Brand | NEB | 7 | Brands Name | ಪೆಪ್ಸಿ, ಕೋಲ (pepsi, kOla) |
| Measure | NEM | 8 | Any measure | ೪,೫೦೦ ರೂ, ೫ ಕೆ.ಜಿ. (4,500 rU, 5 ke.ji.) |
| Number | NEN | 9 | Numeric value | ೩.೧೪, ೪,೫೦೦ (3.14, 4,500) |
| Time | NETI | 10 | date, month, year etc | ೩ನೇ ಸೆಪ್ಟೆಂಬರ್ ೧೯೯೧ (3nE sepTeMbar 1991) |
| Abbreviation | NEA | 11 | Name in short form | ಎನ್ ಎಲ್ ಪಿ, ಬಿ ಜೆ ಪಿ (en el pi, bi je pi) |
| Noun entity | NE | 12 | Other than names (nouns) | ಕತೆಗಾರ (kategAra) |
| Not a NE | NONE | 22 | Not a named entity | ಮಳೆ, ಹೊಗರು, ಹೈ, ಕ (maLe, hOgu, hai, ka) |

The Algorithm Gives The Implementation Procedure Of The MNB Model.

*Algorithm:*
1. Import tools for MNB implementation from python libraries.
2. Read the Development-set of tagged Kannada corpus and divide into N folds (N=10).
3. Reserve one fold of the tagged data as the Dev-test-set.
4. Take all other tagged corpus folds for Training, separate words, tags and store in two lists, x_train, y_train_tag_list. Assign label to each tag and store in another list, y_train.
5. Feature extraction and training of MNB classifier
   Training data: x_train , Y_train from step5
   - *Define vectorizer by the statement*
     **vectorizer = TfidfVectorizer (min_df=1, ngram_range =(1,2), stop_words='english', strip_accents='unicode', norm='l2')**
   - *Read system time t0*
   - *Transform **x_train** to vectors by using **X_train = vectorizer.fit_transform(x_train)***
   - *Define MNB classifier using **clf = MultinomialNB()***
   - *Train MNB classifier using **mnb_classifier = clf.fit (X_train, y_train)***
   - *Read system time now and determine training time,*
     **Train_-time = time () - t0**
   - *Print Feature extraction and Training time of MNB classifier in seconds.*
6. Take reserved fold of the Dev-test-set, separate words, tags and store in two lists, **x_test**, **actual_tag_list.** Assign tag labels and store in another list called **y_test.**
7. validation of MNB model and Feature extraction of Dev-test-set using MNB classifier:
   Dev-test-set : **x_test** : **y_test** from step7
   - Read system time t0
   - Transform **x_test** to vectors by using **X1_test = vectorizer.fit_transform(x_test)**
   - Predict labels of **X_test** by using the statement **mnb_classifier.predict (X1_test)**
   - Read system time now and determine testing time,
     **Fold_Test_time = time () - t0**
8. Attach predicted tags to the Dev-test-set words and print the resultant tagged sequence.
9. Find metrics and print evaluation metrics: Accuracy, Precision, Recall, and F1-measure.
10. Print classification report: individual Precision, Recall, F1_measure and average values.
11. Repeat steps 4 to 11 for all the N folds of Dev-test-sets, and then combine the scores from those evaluations. This technique is known as cross-validation.
12. Take Test-set, tokenize and store in a list called **x_test**, find tags manually and store in a list called **actual_tag_list** and assign tag labels and store in another list called y**_test.**
13. Testing of MNB model and Feature extraction of test data using MNB classifier
    Test data: **x_test** : **y_test** from step11
    - Read system time t0
    - Transform **x_test** to vectors by using **X1_test = vectorizer.fit_transform(x_test)**
    - Predict labels of **X_test** by using the statement **mnb_classifier.predict (X1_test)**
14. Read system time now and determine testing time,
    **Test-time = time () - t0**
15. Attach predicted tags to the Test-set words and print the resultant tagged sequence.
16. Find metrics: Accuracy, Precision, Recall, and F1-measure.
17. Print classification report: individual P, R, F1_mes and average values.
    Plot graphs for the visualization of evaluated results.

## 6. RESULTS AND DISCUSSIONS

The proposed system is designed and implemented as discussed in sections 4 and 5. The system is tested using several test cases, containing training corpus of size 95,170 tokens and test corpus of 5,000 tokens. It is to be noted that the system achieves an average accuracy of 83%. The details of the results obtained are as given below. The system's performance is measured in terms of Precision (P), Recall (R) and F1-measure (F1). The details of the Corpus created in this work are given in section 4. The nature of input Test-set sequence and output tagged sequence are given in Table2 and Table3 respectively. The corpus size and program run time is tabulated in Table4. Table5 shows the results of 10 fold cross validation where validation fold is of size 9,517 tokens. Table 6 Indicates The Final Results Of Test-Set Corpus. Graphical Representation Of MNB Results On Test-Set Corpus Is Plotted As In Figure3.

Table 2. Input Test-Set Sequence.

ಾಹಿಂಗ್ಬನ್ (ಹಿಟಬ): ಮುಂಬರುವ ಲೋಕಸಭಾ ಚುನಾವಣೆ ಬಳಕ ನರೇಂದ್ರ–ಮೋದಿಯೊಂದಿಗೆ. ಅಮೆರಿಕ ರಾಜತಾಂತ್ರಿಕ ಕೆಲಸ ನಿರ್ವಾಹಕರು ಸಿದ್ಧವಿದ್ದು ಇಲ್ಲಿ ವೀಸಾ ಪ್ರಶ್ನೆಯೇ ಇಲ್ಲ ಎಂದು ಅಮೆರಿಕ ಸ್ಪಷ್ಟಪಡಿಸಿದೆ. ………..

Table 3. Output Tagged Sequence.

ಾಹಿಂಗ್ಬನ್/NEL (ಹಿಟಬ):/NONE ಮುಂಬರುವ/NONE ಲೋಕಸಭಾ/NE ಚುನಾವಣೆ/NE ಬಳಕ/NONE ನರೇಂದ್ರ/NEPB ಮೋದಿಯೊಂದಿಗೆ./NEPE ಅಮೆರಿಕ/NEL ರಾಜತಾಂತ್ರಿಕ/NONE ಕೆಲಸ/NONE ನಿರ್ವಾಹಕರು/NONE ಸಿದ್ಧವಿದ್ದು/NONE ಇಲ್ಲಿ/NONE ವೀಸಾ/NONE ಪ್ರಶ್ನೆಯೇ/NONE ಇಲ್ಲ/NONE ಎಂದು/NONE ಅಮೆರಿಕ/NEL ಸ್ಪಷ್ಟಪಡಿಸಿದೆ./NONE ………..

Table 4. Corpus size and program Run time.

| | | |
|---|---|---|
| The training set size for the Model | : | 95,170 words |
| Total number of samples treated by the classifier | : | 95,170 words |
| Total number of features extracted by the classifier | : | 33,269 (vocabulary words) |
| Feature extraction Time (Training of MNB model) | : | 7.407 sec |
| The test set size for the Model | : | 5,000 words |
| Feature extraction Time for test data | : | 2.765 sec |

Table 5. Results of 10 Fold Cross Validation.

| FOLDS | Precision | Recall | F1 - score | Support |
|---|---|---|---|---|
| 1 | 0.79 | 0.78 | 0.78 | 9517 |
| 2 | 0.63 | 0.62 | 0.61 | 9517 |
| 3 | 0.70 | 0.67 | 0.67 | 9517 |
| 4 | 0.76 | 0.74 | 0.73 | 9517 |
| 5 | 0.82 | 0.81 | 0.81 | 9517 |
| 6 | 0.82 | 0.82 | 0.82 | 9517 |
| 7 | 0.83 | 0.83 | 0.83 | 9517 |
| 8 | 0.81 | 0.81 | 0.81 | 9517 |

| | | | | | |
|---|---|---|---|---|---|
| 9 | | | 0.83 | 0.83 | 0.83 | 9517 |
| 10 | | | 0.87 | 0.81 | 0.83 | 9517 |
| **Average / Total** | | | 78.6 % | 77.2% | 77.2% | 95170 |

Table 6. Result of Test Corpus.

| **Named Entity (NE)** | **Tag** | **Tag label** | **Precision** | **Recall** | **F1 - score** | **Support** |
|---|---|---|---|---|---|---|
| Person | NEP | 0 | 0.62 | 0.59 | 0.60 | 229 |
| | NEPB | 13 | 0.33 | 0.83 | 0.48 | 6 |
| | NEPI | 14 | 0.00 | 0.00 | 0.00 | 0 |
| | NEPE | 15 | 0.38 | 0.46 | 0.41 | 26 |
| Location | NEL | 1 | 0.47 | 0.76 | 0.58 | 66 |
| | NELB | 16 | 0.67 | 0.80 | 0.73 | 5 |
| | NELI | 17 | 0.00 | 0.00 | 0.00 | 0 |
| | NELE | 18 | 0.50 | 0.50 | 0.50 | 4 |
| Organization | NEO | 2 | 0.25 | 0.67 | 0.36 | 9 |
| | NEOB | 19 | 0.00 | 0.00 | 0.00 | 0 |
| | NEOI | 20 | 0.00 | 0.00 | 0.00 | 0 |
| | NEOE | 21 | 0.00 | 0.00 | 0.00 | 1 |
| Designation | NED | 3 | 0.39 | 0.86 | 0.54 | 21 |
| Term | NETE | 4 | 0.25 | 0.39 | 0.30 | 54 |
| Title-Person | NETP | 5 | 0.42 | 0.50 | 0.46 | 22 |
| Title-Object | NETO | 6 | 0.44 | 0.39 | 0.42 | 56 |
| Brand | NEB | 7 | 0.00 | 0.00 | 0.00 | 0 |
| Measure | NEM | 8 | 0.47 | 0.59 | 0.52 | 56 |
| Number | NEN | 9 | 0.52 | 0.39 | 0.45 | 116 |
| Time | NETI | 10 | 0.24 | 0.48 | 0.32 | 23 |
| Abbreviation | NEA | 11 | 0.05 | 1.00 | 0.10 | 1 |
| Noun entity | NE | 12 | 0.49 | 0.72 | 0.58 | 309 |
| Not a NE | NONE | 22 | 0.92 | 0.84 | 0.88 | 3996 |
| Average / Total | | | 0.83 | 0.79 | 0.81 | 5000 |

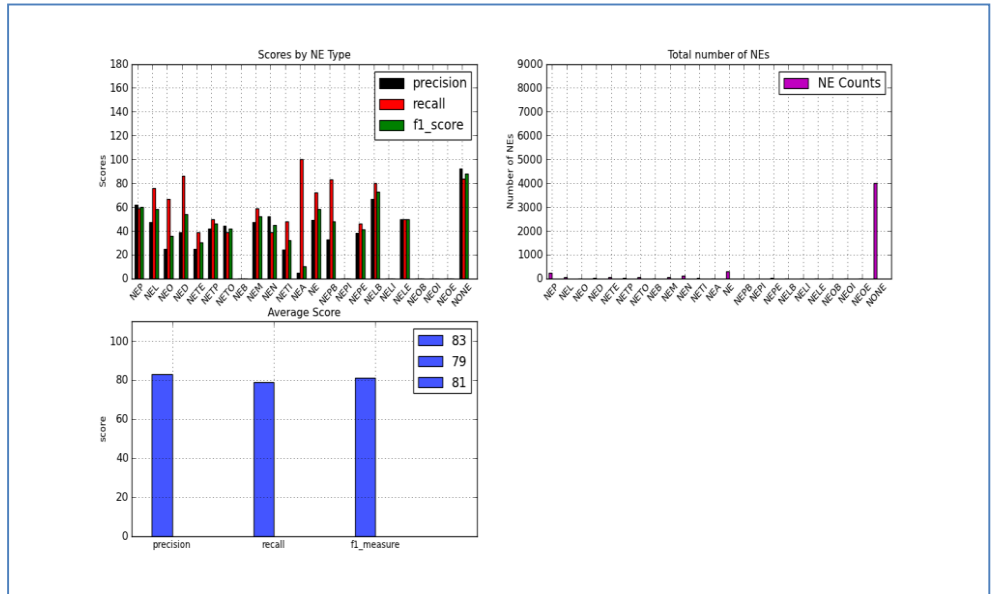

Figure 3. Graphical representation of MNB results on Test Corpus

## 7. CONCLUSION

Natural Language Processing is an important research area containing challenging issues to be investigated. NERC is a class of NLP which is used for extracting named entities from unstructured data. In this context this paper focuses on NERC in Kannada Language as it is found that little work is done in this area. In this direction, we have conducted a vast survey in the related area of NLP and based on the survey we proposed a problem and the methodology that has been formulated. Various Modeling techniques are investigated, out of which a design of Supervised MNB is done. The results obtained are recorded and evaluated. We developed an efficient model and trained on a Corpus consisting of around 95170 words. From this training Corpus, some test samples are chosen and fed as input to the MNB. It is interesting to note that the model recognizes the named entities with an average F-measure of 81% and 10 fold cross validation F-measure of 77.2%.

## References

[1] Elizabeth D Liddy. *Natural Language Processing. In Encyclopedia of Library and Information Science*. 2nd edition, 2001.

[2] James Allen. Natural Language Understanding. Pearson Publication Inc., 2nd edition, 2007.

[3] Kavi Narayana Murthy. *Natural Language Processing*. Ess Ess Publications for Sarada Ranganathan Endowment for Library Science, Bangalore, INDIA, 1st edition, 2006.

[4] Gobinda G. Chowdhury. Natural language processing, annual review of information science and technology. 37(1):51–89, 2003.

[5] Yassine Benajiba, Mona T Diab, and Paolo Rosso. Using language independent and language specific features to enhance Arabic named entity recognition. *Int. Arab J. Inf. Technol.*, 6(5):463–471, 2009.

[6] Padmaja Sharma, Utpal Sharma, and Jugal Kalita. The first steps towards Assamese named entity recognition. *Brisbane Convention Center*, 1:1–11, 2010.

[7] Asif Ekbal Rejwanul Haqueand and Sivaji Bandyopadhyay. Named entity recognition in Bengali. A conditional random field approach. Pages 589–594, 2008.

[8] Asif Ekbal and Sivaji Bandyopadhyay. Bengali named entity recognition using support vector machine. In *IJCNLP*, pages 51–58, 2008.

[9] Asif Ekbal and Sivaji Bandyopadhyay. Development of Bengali named entity tagged corpus and its use in NER systems. In *IJCNLP*, pages 1–8, 2008.

[10] Maksim Tkachenko, Andrey Simanovsky, and St Petersburg. Named entity recognition: Exploring features. In *Proceedings of KONVENS*, pages 118–127, 2012.

[11] Alireza Mansouri, Lilly Suriani Affendey, and Ali Mamat. Named entity recognition approaches. *International Journal of Computer Science and Network Security*, 8:339–344, 2008.

[12] David Nadeau, Peter Turney, and Stan Matwin. Unsupervised named-entity recognition: Generating gazetteers and resolving ambiguity. *Published at the 19th Canadian Conference on Artificial Intelligence*, June 2006.

[13] Monica Marrero, Sonia Sanchez-Cuadrado, Jorge Morato Lara, and George Andreadakis. Evaluation of named entity extraction systems. *Advances in Computational Linguistics, Research in Computing Science*, 41:47–58, 2009.

[14] Sujan Kumar Saha, Partha Sarathi Ghosh, Sudeshna Sarkar, and Pabitra Mitra. Named entity recognition in Hindi using maximum entropy and transliteration. *Research journal on Computer Science and Computer Engineering with Applications*, pages 33–41, 2008.

[15] Li Wei and McCallum Andrew. Rapid development of hindi named entity recognition using conditional random fields and feature induction (short paper). *ACM Transactions on Computational Logic*, 2004.

[16] Sujan Kumar Saha, Sudeshna Sarkar, and Pabitra Mitra. A hybrid feature set based maximum entropy Hindi named entity recognition. In *IJCNLP*, pages 343–349, 2008.

[17] Sudha Morwal and Nusrat Jahan. Named entity recognition using hidden markov model (hmm): An experimental result on Hindi, urdu and marathi languages. *International Journal of Advanced Research in Computer Science and Software Engineering (IJARCSSE)*, 3(4):671–675, 2013.

[18] Deepti Chopra and Sudha Morwal. Detection and categorization of named entities in Indian languages using Hidden Markov Model. *International Journal of Computer Science*, 2013.

[19] Nusrat Jahan, Sudha Morwal, and Deepti Chopra. Named entity recognition in Indian languages using gazetteer method and hidden markov model: A hybrid approach. *IJCSET, March*, 2012.

[20] S Biswas, MK Mishra, S Acharya Sitanath, and S Mohanty. A two stage language independent named entity recognition for indian languages. *IJCSIT International Journal of Computer Science and Information Technologies*, 1(4):285–289, 2010.

[21] Animesh Nayan, B Ravi Kiran Rao, Pawandeep Singh, Sudip Sanyal, and Ratna Sanyal. Named entity recognition for Indian languages. In *IJCNLP*, pages 97–104, 2008.

[22] Sujan Kumar Saha, Sanjay Chatterji, Sandipan Dandapat, Sudeshna Sarkar, and Pabitra Mitra. A hybrid approach for named entity recognition in Indian languages. *NER for South and South East Asian Languages*, pages 17–24, 2008.

[23] Erik F Tjong Kim Sang and Fien De Meulder. Introduction to the conll-2003 shared task: Language-independent named entity recognition. In *Proceedings of the seventh conference on*

*Natural language learning at HLT-NAACL 2003-Volume 4*, pages 142–147. Association for Computational Linguistics, 2003.

[24] Asif Ekbal and Sivaji Bandyopadhyay. Named entity recognition using support vector machine: A language independent approach. *International Journal of Electrical, Computer, and Systems Engineering*, 4(2):155–170, 2010.

[25] Asif Ekbal, Rejwanul Haque, Amitava Das, Venkateswarlu Poka, and Sivaji Bandyopadhyay. Language independent named entity recognition in indian languages. In *IJCNLP*, pages 33–40, 2008.

[26] Kishorjit Nongmeikapam, Tontang Shangkhunem, Ngariyanbam Mayekleima Chanu, Laishram Newton Singh, Bishworjit Salam, and Sivaji Bandyopadhyay. Crf based name entity recognition (ner) in manipuri: A highly agglutinative indian language. In *Emerging Trends and Applications in Computer Science (NCETACS), 2011 2nd National Conference on*, pages 1–6. IEEE, 2011.

[27] Thoudam Doren Singh, Kishorjit Nongmeikapam, Asif Ekbal, and Sivaji Bandyopadhyay. Named entity recognition for manipuri using support vector machine. In *PACLIC*, pages 811–818, 2009.

[28] Sitanath Biswas, SP Mishra, S Acharya, and S Mohanty. A hybrid oriya named entity recognition system: harnessing the power of rule. *International Journal of Artificial Intelligence and Expert Systems (IJAE)*, 1(1):1–6, 2010.

[29] Vishal Gupta and Gurpreet Singh Lehal. Named entity recognition for Punjabi language text summarization. *International Journal of Computer Applications*, 33(3):28–32, 2011.

[30] S Pandian, Krishnan Aravind Pavithra, and T Geetha. Hybrid three-stage named entity recognizer for Tamil. *INFOS*, 2008.

[31] G.V.S.Raju B.Srinivasu Dr.S.Viswanadha Raju K.S.M.V.Kumar. Kumar, named entity recognition for Telugu using maximum entropy model. *Journal of Theoretical and Applied Information Technology (JATIT)*, 13:125–130, 2010.

[32] Kommaluri Vijayanand and RP Seenivasan. Named entity recognition and transliteration for Telugu language. *Parsing in Indian Languages*, Special Volume: Problems of Parsing in Indian Languages: 64–70, may 2011.

[33] P Srikanth and Kavi Narayana Murthy. Named entity recognition for Telugu. In *IJCNLP*, pages 41–50, 2008.

[34] Praneeth Shishtla, Karthik Gali, Prasad Pingali, and Vasudeva Varma. Experiments in Telugu NER: A conditional random field approach. In *IJCNLP*, pages 105–110, 2008.

[35] BR Shambhavi, Kumar P Ramakanth, and G Revanth. A maximum entropy approach to Kannada part of speech tagging. *International Journal of Computer Applications*, 41(13):9–12, 2012.

[36] Ramasamy Veerappan, PJ Antony, S Saravanan, and KP Soman. A rule based Kannada morphological analyzer and generator using finite state transducer. *Proceedings of International Journal of Computer Applications (0975-8887)*, 27(10), 2011

[37] S Amarappa and SV Sathyanarayana. Named entity recognition and classification in Kannada language. *International Journal of Electronics and Computer Science Engineering*, 2(1):281–289, Jan 2013.